\documentclass[11pt, a4paper, logo, singlecolumn, copyright]{gensyn}

\usepackage{booktabs}
\usepackage{hyperref}
\usepackage{url}
\usepackage{alertmessage}
\usepackage{xspace}
\usepackage{graphicx}
\usepackage{multirow}
\usepackage{rotating}
\usepackage{tikz,lipsum}

\usepackage{amsmath}
\usepackage{amssymb}
\usepackage{mathtools}
\usepackage{amsthm}
\usepackage{colortbl}
\usepackage{placeins}
\usepackage{subcaption}
\usepackage{adjustbox}
\usepackage{wrapfig}

\usepackage[colorinlistoftodos,textsize=tiny,textwidth=35pt]{todonotes}
\newcommand{\Comments}{1}
\newcommand{\mynote}[2]{\ifnum\Comments=1\textcolor{#1}{#2}\fi}
\newcommand{\mytodo}[2]{\ifnum\Comments=1\todo[linecolor=#1!80!black,backgroundcolor=#1,bordercolor=#1!80!black]{#2}\fi}

\ifnum\Comments=1
\paperwidth=\dimexpr \paperwidth + 50pt\relax
\oddsidemargin=\dimexpr\oddsidemargin + 25pt\relax
\evensidemargin=\dimexpr\evensidemargin + 25pt\relax
\marginparwidth=\dimexpr \marginparwidth + 25pt\relax
\fi

\usepackage{array}
\newcolumntype{P}{>{\centering\arraybackslash}p{2.5cm}}
\newcolumntype{M}{>{\centering\arraybackslash\footnotesize}m{.78cm}}
\newcolumntype{S}{>{\centering\arraybackslash\tiny}m{2cm}}

\newtcbtheorem[auto counter,number within=section]{obs}%
{Observation}{fonttitle=\bfseries\upshape, fontupper=\slshape,
	arc=0mm, colback=cyan!5!white,colframe=cyan!75!white}{theorem}

\newtcbtheorem[auto counter,number within=section]{ins}%
{Insight}{fonttitle=\bfseries\upshape, fontupper=\slshape,
	arc=0mm, colback=green!5!white,colframe=green!75!white}{theorem}

\usepackage{amsmath,amsfonts,bm}

\def\eqref#1{equation~\ref{#1}}

\def\1{\bm{1}}

\DeclareMathAlphabet{\mathsfit}{\encodingdefault}{\sfdefault}{m}{sl}
\SetMathAlphabet{\mathsfit}{bold}{\encodingdefault}{\sfdefault}{bx}{n}

\newcommand{\R}{\mathbb{R}}

\DeclareMathOperator*{\argmax}{arg\,max}

\title{IR3DE: A Linear Router for Large Language Models}

\author[1]{Eros Fanì}
\author[1]{O\u{g}uzhan Ersoy}

\affil[1]{Gensyn}

\correspondingauthor{eros@gensyn.ai}

\reportnumber{} %

\definecolor{darkgreen}{rgb}{0,0.7,0}

\newcommand\norm[1]{\left\lVert#1\right\rVert}
\newcommand\method{\textsc{IR3DE}\xspace}

\newcommand{\eg}{\textit{e.g.,}\xspace}
\newcommand{\ie}{\textit{i.e.,}\xspace}

\newcommand{\clm}{\textsc{CLM}\xspace}
\newcommand{\clmII}{\textsc{CLMlarge}\xspace}
\newcommand{\reasoning}{\textsc{Reasoning}\xspace}

\begin{abstract}
	\vspace{1em}
Foundational Large Language Models (LLMs) demonstrate proficiency on a wide range of general tasks, and achieve remarkable results on various specialized tasks via domain-expert LLMs.
With the ever-growing list of available LLMs, inference routers are being proposed to select the most appropriate LLM for each prompt.
However, existing routing methods either optimize cost across weak-to-strong generalist LLMs or require substantial training to support domain-expertise routing.
In this paper, we propose \method, a Ridge Regression-based Router for Domain Experts that provides cheap and fast routing decisions for each prompt. 
We evaluate \method in two Causal Language Modeling (CLM) settings where the tasks are next-token prediction for all domains, and one reasoning setting where each domain has its own distinct reasoning task.
Despite being a linear router, \method achieves performance comparable to the other baselines in both CLM settings, and surpassing them in the reasoning setting, with a normalized performance of 98.4\%. 
Moreover, \method enables the addition or removal of new domain experts without requiring the router to be retrained from scratch, allowing a dynamic set of LLMs to be served with minimal disruption to the router itself. Our code is available at: \url{github.com/gensyn-ai/IR3DE}.

\end{abstract}

\begin{document}

\maketitle

\section{Introduction}
\label{sec:intro}

Large Language Models (LLMs) have shown remarkable performance gains on both language and cognitive tasks in recent years.
Foundation (or generalized) models perform well across a wide range of tasks~\citep{radford2019language,touvron2023llama,qwen2023}, whereas expert (or specialized) ones excel on specific tasks such as code generation~\citep{chen2021evaluating}, mathematical problem-solving~\citep{cobbe2021training}, and instruction following~\citep{zhou2023instruction}.
Depending on a user's needs or the prompt, different models can be used. 
Routing all queries to a generalist model can be suboptimal because such models may lack the necessary domain knowledge or be unnecessarily expensive.
Thus, it is beneficial to route each query to the most relevant (expert) model.

LLM routing allows us to dynamically assign each query to the most appropriate model based on selection criteria.
A common selection criterion is the cost-performance trade-off between weak and strong LLMs, where the models have similar capabilities but differ in capacity~\citep {routerLLM,graphRouter,irt-router,uniRoute}. 
Here, the router makes the selection mainly based on the difficulty of the query, i.e., sending difficult queries to the strong (but large) model and vice versa, rather than on their expertise.
There are also routers that assign queries based on each LLM's expertise, aiming to maximize accuracy~\citep{modem_router,polyrouter}. 
However, such expert routers utilize additional (language) models either to classify each query or to obtain token embeddings using the last hidden layer representations.
Therefore, they require collecting domain datasets to train the router, which may not be feasible due to privacy concerns, or they use a language model as the router. %

In this paper, we propose \method, which enables efficient routing to the most suitable domain expert.
\method utilizes a closed-form solution via ridge regression, inspired by its use in federated learning~\citep{afonin2021towards,cai2022efficient, huang2022coresets, fani2024accelerating} and MoE routers~\citep{fani2026dume}. 
We use a linear ridge-regression token router and select the expert LLM based on the confidence of the top-k tokens.
Since the domain statistics can be computed asynchronously for the solution, \method does not require collecting the datasets in a single location. Moreover, the cost of running \method is irrelevant, as it only requires token embedding with any desired embedding layer and the inversion of a small matrix (typically about a 1k-by-1k matrix) to be performed only one time.

Our contributions can be summarized as follows:
\begin{itemize}
    \item We introduce \method, a linear expert router for LLMs. \method router consists of two components: a \textit{token router} (TR) based on ridge regression and a \textit{sample route selector} (SRS). Thanks to its linear construction, \method is cheaper and faster than LM-based baselines. Moreover, it enables the addition or removal of new domain experts without requiring the router to be retrained from scratch.
    \item We evaluate \method in two CLM settings and one reasoning setting, across several domains and tasks. \method achieves performance comparable to the other baselines in both CLM settings, and surpasses them in the reasoning setting.
    \item We present three variants of \method (specifically, of the SRS) based on averaging, majority voting, or entropy of TR outputs. Experimental results show that the entropy-based approach achieves the best performance on complex reasoning tasks, where precision becomes more crucial.
\end{itemize}

\section{Related Work}
\label{sec:relatedwork}

With the increasing number of LLMs, including both generalist and expert models, routing has become increasingly important for dynamically assigning each query to the most appropriate model. 
Early work on LLM routing primarily focused on the cost–quality trade-off in fixed pools of generalist models~\citep{hybrid_llm,routerLLM,irt-router}. 
By estimating prompt difficulty, these methods route queries to an adequate model rather than always invoking the most powerful (and costly) one. 
In addition to a single LLM selection, cascading approaches have been proposed in~\citep{frugalgpt2024chen,disrouter2025zheng,cascading2025dekoninck},  where first a weaker LLM generates the output and depending on the quality or confidence, it is escalated to a stronger LLM. 
Subsequent work has broadened this paradigm by taking into account the reasoning strategies and task profiles~\citep{route_to_reason, r2_router, liu2026task}, and by enabling routers to accommodate previously unseen models without retraining from scratch~\citep{graphRouter,uniRoute,icl_router}.

Another selection criterion for routing decisions is accuracy. In accuracy-oriented routing, where the router assesses the domain of a prompt and forwards it to the most relevant (specialist) model ~\citep{modem_router,polyrouter}.
Finally, there are extensions using multiple experts: 
Symbolic-MoE~\citep{symbolicMoE} selects several experts for a given query and combines their outputs using an aggregator, and HierRouter~\citep{hierRouter} proposes a hierarchical routing approach that selects expert models in a multi-hop inference setting.

The most relevant works are \emph{MoDEM}~\citep{modem_router} and \emph{PolyRouter}~\citep{polyrouter}.
The MoDEM router uses a DeBERTa v3 model \citep{he2021debertav3} trained on the union of all domain datasets used to train the domain experts.
Because of this, this method is unsuitable in cases where it is not possible to collect data from all domains on a single node due to privacy constraints or limited communication budget, or when the computational or memory budget is insufficient for additional training. 
PolyRouter presents multiple routing methods: (i) \emph{BERT-router} that trains a BERT model \citep{devlin2019bert} to classify the domains; (ii) \emph{MLP-router} that trains a 2-layer perceptron using a Bag-of-Words representations of training queries; and (iii) \emph{1NN-router}, which uses sentence-transformer embeddings to find the nearest training query and select the corresponding expert.
All these methods require additional (language) models either to classify each query or to obtain token embeddings.

\section{Method}
\label{sec:method}

\begin{wrapfigure}{r}{.4\textwidth}
    \centering
    \includegraphics[width=\linewidth]{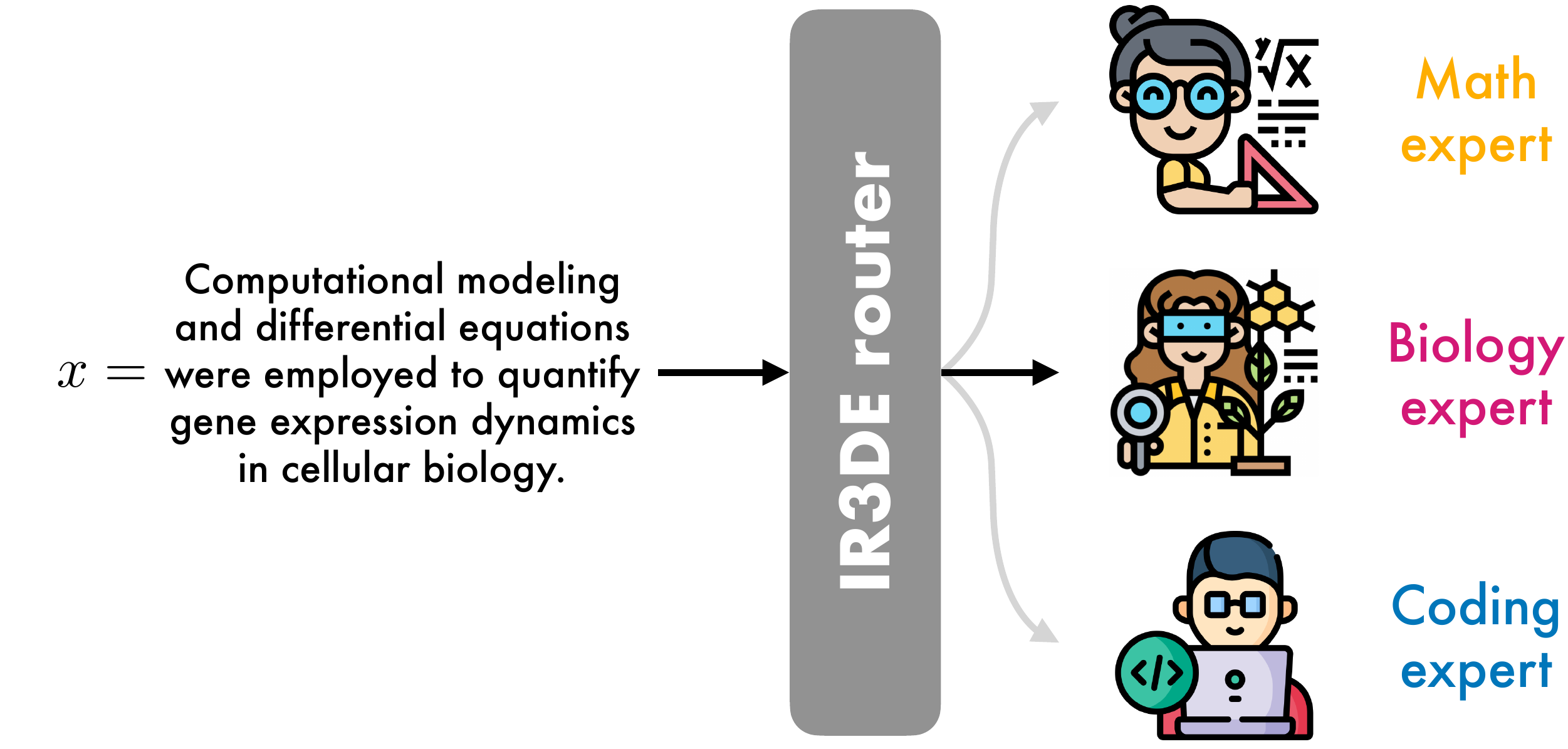}
    \caption{Given an input text $x$, our \method router selects the most appropriate expert solely based on the token embeddings.}
    \label{fig:inferencerouter}
\end{wrapfigure}

We have access to a set of LM experts $\{f_d\}_{d=1}^D$. Each expert has been trained on a different domain dataset  %
$\mathcal{D}_d$, $d=1 \dots D$ for some upstream or downstream task (\eg math, biology, coding, ...), whose samples belong to different distributions. Our objective is to construct a routing mechanism that, given an input text, is able to select the expert who would maximize %
the performance for that input. Performance could be any desired metric, depending on the tasks for which the experts have been trained. For instance, for causal language modeling, the desired metric could be perplexity, while for coding it could be pass@1. %
The setting is represented in \Cref{fig:inferencerouter}.

\begin{figure*}[ht!]
    \centering
    \begin{minipage}{0.46\linewidth}
        \centering
        \includegraphics[width=\linewidth]{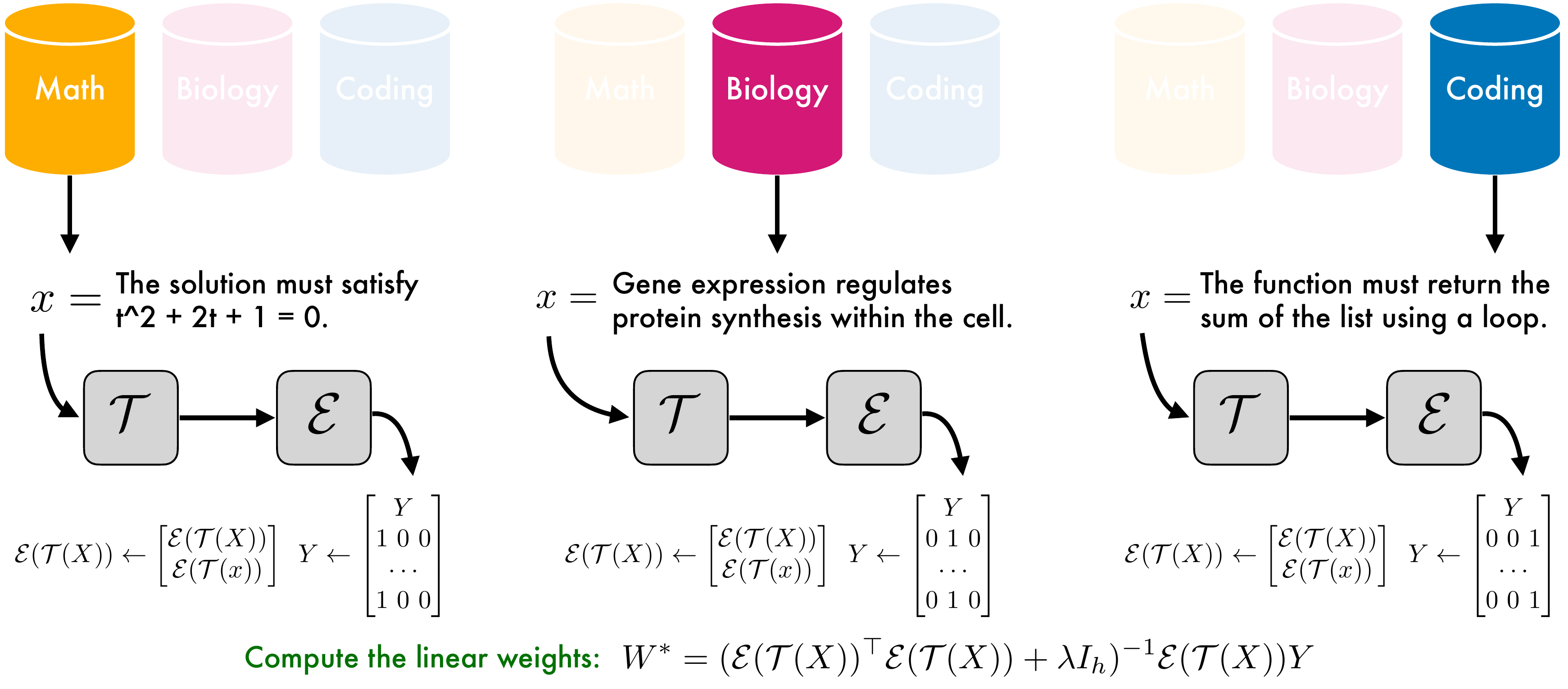}
        \subcaption{Token Router.}
        \label{fig:tr}
    \end{minipage}
    \hfill
    \begin{minipage}{0.46\linewidth}
        \centering
        \includegraphics[width=\linewidth]{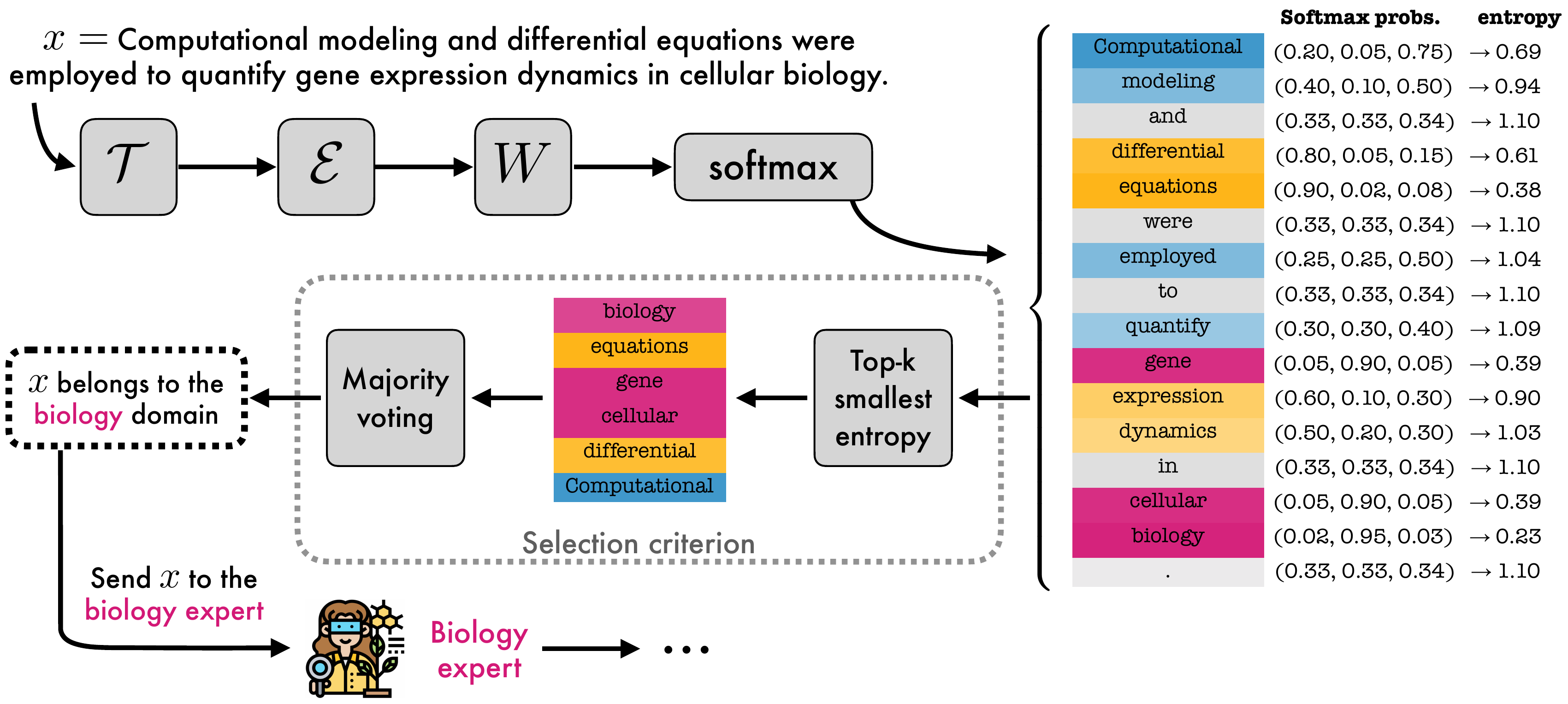}
        \subcaption{Sample Route Selector.%
        }
        \label{fig:srs}
    \end{minipage}

    \caption{%
    \textbf{(a)} Construction of the Token Router. Input samples are forwarded to the tokenizer and the embedding layer. The embeddings and one-hot encoding vectors are stacked together and eventually used to compute the optimal RLS weights. \textbf{(b)} At inference time, the input samples are forwarded to the token router to extract softmax probabilities associated with each token. Then, entropy is computed for the softmax probabilities of each token, and only the top-k tokens with the smallest entropy are retained (In the image, k=6). Finally, each of the surviving tokens cast a vote to elect the final expert to whom $x$ will be forwarded.}
    \label{fig:trandsrs}
\end{figure*}

Our \method router is constituted of two components: a Token Router and a Sample Route Selector. We provide an overview of our \method method in \Cref{fig:trandsrs}.

\subsection{Token Router (TR)}

Given an input $x$, we first construct the Token Router $\mathcal{R}$:  
\begin{equation*}
    \mathcal{R}(x) = softmax(\mathcal{E}(\mathcal{T}(x)) W).
\end{equation*}

Without loss of generality, assume that the input text $x$ is composed of $T$ tokens. The tokenizer $\mathcal{T}$ returns a list of token ids, which are then forwarded to a pre-trained embedding layer $\mathcal{E}$ to construct a matrix of embeddings $\mathcal{E}(\mathcal{T}(x)) \in \R^{T \times h}$. Finally, linear weights $W \in \R^{h \times C}$ are multiplied to the matrix of embeddings, to construct softmax probabilities for each token. Therefore, each input token will be associated to a softmax probability vector $\mathcal{R}(x)_t$. Please note that any choice for the tokenizer $\mathcal{T}$ and embedding layer $\mathcal{E}$ works, given that the $\mathcal{E}$ has been trained using the tokenizer $\mathcal{T}$. In other words, their choice is independent from the tokenizers and embedding layers actually used by the experts, which could also vary.

Now, let $\mathcal{E}(\mathcal{T}(X)) \in \R^{n \times C}$ be the matrix of stacked embeddings of all the tokens of each sample of each domain dataset, and $Y \in \R^{n \times C}$ be the matrix of stacked one-hot vectors associated with each sample, such that $Y_{ij} = 1$ iff the original sample of token $i$ belongs to domain $j$, $0$ otherwise. We obtain the linear weights $W$ by solving the regularized least squares problem $\min_{W \in R^{h \times C}} \left [ \norm{\mathcal{E}(\mathcal{T}(X)) - Y}^2 + \lambda \norm{W}^2 \right ]$, which admits the following closed form solution:
\begin{equation*}
    W^* = (\mathcal{E}(\mathcal{T}(X))^\top \mathcal{E}(\mathcal{T}(X)) + \lambda I_h)^{-1} \mathcal{E}(\mathcal{T}(X)) Y,
\end{equation*}
where $\lambda \in \R^+$ controls Tikhonov regularization, and $I_h$ is the identity matrix of size $h \times h$. To avoid large matrix multiplications, it is possible to collect RLS statistics batch-wise by exploiting the properties of the closed-form RLS solution. For instance, for batches $j$ of $J$ tokens, we can equivalently compute:
\begin{align*}
    &A \coloneqq \sum_j \left [\mathcal{E}(\mathcal{T}(X)) \right ]_{j:j+J}^\top \left [\mathcal{E}(\mathcal{T}(X))\right ]_{j:j+J}, \\
    &B \coloneqq \sum_j \left [\mathcal{E}(\mathcal{T}(X)) \right ]_{j:j+J}^\top \left [Y \right ]_{j:j+J}, \\
    &W^* = (A + \lambda I_h)^{-1}B.
\end{align*}

This formulation also demonstrates that our solution is suitable for decentralized settings where domain datasets cannot be easily collected on a single node, since each dataset can be treated as a separate batch (or multiple separate batches). Moreover, it allows new expert models to join at any time without having to reconstruct the router from scratch. Alternatively, it is also possible to train $\mathcal{R}$ with any desired loss function, \eg using the cross-entropy loss, and using $\mathcal{E}(\mathcal{T}(X))$ and $Y$ as the dataset.

\subsection{Sample Route Selector (SRS)}
Given an input text $x$, the Token Router returns a matrix of $T$ softmax probabilities $s = [s_1^\top, s_2^\top, \dots, s_T^\top]^\top$, where $s_t = \mathcal{R}(x)_t$ for each token $t$ of $x$. Then, the Sample Route Selector computes the entropy vector $e \in \R^T$, such that $e_t = -\sum_{d=1}^D s_{td} \log s_{td}$ is the Shannon entropy of $s_t$. SRS then selects the tokens with the $\min(k, T)$ smallest entropy in $e$, returning $\hat{s} \in \R^{\min(k, T)}$. Finally, SRS computes the $\argmax$ of each row of $\hat{s}$, returning the \emph{vote} cast by each token of $x$ of where to route the input text $x$, and the final expert to which $x$ is routed is selected via majority voting.

Motivations on the Sample Route Selector construction are provided in \Cref{sec:kentropy}.

\paragraph{Variants.} In addition to our chosen SRS, we also present two additional variants:
\begin{itemize}
\item \emph{\method-all}: in this variant, instead of top-k, we use all tokens in each prompt (and the number of tokens per prompt is capped at 1024 in our experiments). Therefore, in this scenario, there is no entropy-based filtering, and all tokens contribute to selecting the final expert via majority voting. In \Cref{sec:kentropy}, we explain how this can introduce noise into the majority-voting process.
\item \emph{\method-avg}: in this variant, we modify our SRS by first computing the average of the token embeddings and selecting the final domain with the argmax of the softmax probabilities computed on the average token embedding vector. This is a cheap variant that does not require the SRS but could hinder the single tokens' ability to determine the correct routing domain, since a strong signal compression is applied.
\end{itemize}

\section{Experiments}
\label{sec:exp}

\subsection{Setup}

\begin{table*}[t]
    \caption{Datasets used for evaluating the reasoning experiments, and their evaluation metrics.}
    \label{tab:postdatasets}
    \centering
    \small
    \begin{adjustbox}{width=\linewidth}
        \begin{tabular}{p{2cm} p{9cm} p{5cm}}
            \toprule
            Dataset & Description & Metric \\
            \midrule
            HumanEval \citep{chen2021evaluating} 
            & Code generation benchmark consisting of programming tasks specified by natural-language prompts and associated unit tests.  
            & pass@1: fraction of tasks solved correctly by the first generated completion. \\
            GSM8k \citep{cobbe2021training} 
            & Grade-school mathematics benchmark containing linguistically diverse word problems that require multi-step reasoning. 
            & Accuracy: fraction of examples for which the final answer is correct. \\
            
            M\_ARC \citep{lai2023okapi} 
            & Multilingual version of the ARC challenge benchmark, consisting of multiple-choice question-answering tasks that test reasoning ability. 
            & Accuracy: percentage of questions answered correctly. \\
            
            IFEval \citep{zhou2023instruction} 
            & Instruction-following benchmark for LLMs, with prompts containing verifiable constraints that allow automatic evaluation of compliance. 
            & Task compliance rate: percentage of responses that satisfy all instructions. \\
            \bottomrule
        \end{tabular}
    \end{adjustbox}
\end{table*}

We conduct experiments in three settings: two \emph{Causal Language Modeling} settings, \clm and \clmII, where the task is next-token prediction for all domains, and one \emph{Reasoning} setting (\reasoning), where each domain has its own distinct reasoning task. Below, we present the details of our settings and the baselines compared with our method. All experiments run on an NVIDIA H100 GPU with 80 GB of HBM3 memory.

\paragraph{Expert Models and Domain Datasets.} In the \clm setting, we train our expert models starting from a shared base model. Both the base and expert models are trained using the hyperparameters and datasets outlined in~\citep{ersoy2025hdee}. Initially, we pre-trained a 115M Llama3 seed model \citep{grattafiori2024llama} on the OpenWebText corpus dataset \citep{Gokaslan2019OpenWeb}, then finetuned the expert models on M2D2 domains \citep{reid2022m2d2} (coding, mathematics, physics, history and events, and philosophy and thinking). For \clmII experiments, we finetuned expert models using a 1B Llama3 base model on these domains: mathematics (OpenWebMath \citep{paster2023openwebmath}), biology (peS2o \citep{peS2o}), legal (Pile of Law \citep{hendersonkrass2022pileoflaw}), and dialogue (UltraChat 200k \citep{ding2023enhancing}). In both CLM settings, the metric used for evaluation is perplexity.

For the \reasoning experiments, we use domain-specific LLama3-3B experts from MergeBench \citep{he2025mergebench} on the following domains: coding, mathematics, multilingual understanding, and instruction following. We use the following downstream task datasets for the evaluation: HumanEval \citep{chen2021evaluating} for coding, GSM8k \citep{cobbe2021training} for mathematics, M\_ARC \citep{lai2023okapi} for multilingual understanding, and IFEval \citep{zhou2023instruction} for instruction following. Details on the evaluation metrics for each \reasoning dataset can be found in Table~\ref{tab:postdatasets}. %

Following the same practice as in \cite{fani2026dume}, we report the final results using normalized metrics relative to the performance each domain expert achieves in their respective domain (across all settings, the higher, the better).

\paragraph{Baselines.} We compare our method with the following baselines:
\begin{itemize}
    \item \emph{``Domain'' expert}: as in \cite{hu2024routerbench, polyrouter}, we compare our proposed method against domain experts' performance across all domains.
    \item \emph{Experts average}: similarly, we compare our method against the average of the expert models.
    \item \emph{Random routing}: also used by \citep{polyrouter} as a lower bound, the expert is randomly selected for any given prompt.
    \item \emph{MoDEM}: proposed by \cite{modem_router}, it uses a DeBERTa v3 model \citep{he2021debertav3} trained on the domain datasets. We present two versions of this baseline: \emph{MoDEM-small}, which uses a DeBERTa v3 small model with 44M parameters, and \emph{MoDEM-large}, which uses a DeBERTa v3 large model with 304M parameters. 
    \item \emph{1NN router}: proposed by \citep{polyrouter}, it uses the BERT model \citep{devlin2019bert} to extract embeddings for each input prompt. Then, at inference time, the test samples are embedded using the same BERT model, and each test sample's embedding is compared with all training embeddings using cosine similarity. Finally, the nearest training embedding is selected, and the test sample is forwarded to the expert with the corresponding domain. 
    \item \emph{kNN router}: expands the \emph{1NN router} strategy by selecting $k$ experts, and then letting them elect the winning domain via majority voting. Like the \emph{1NN router}, it uses a BERT model to extract the embeddings.
\end{itemize}

Note that we omit the two other baselines proposed in~\citep{polyrouter}, namely \emph{BERT-router} and \emph{MLP-router}, given that \emph{MoDEM} provides a similar solution using a language model, DeBERTa v3, that is supposed to perform better than a BERT model. Additionally, please observe that \emph{MoDEM-large} router is even larger than the experts in the \clm setting, making this baseline too expensive and impractical for real-world deployment.

For the \emph{kNN router}, in all our settings, we tried all $\text{k} \in {1, 5, 10}$. Moreover, for our \method, we tried all $k \in {1, 2, 5, 10, 20, 50, 100, 200, 500}$ for our top-k entropy-based selection rule.\footnote{We use ``k'' for \emph{kNN router} and ``$k$'' for \method, to emphasize that the two are used in different contexts and with different meanings.} For both \emph{kNN router} and \method, in the following tables, we show only the best-performing result.

\paragraph{Metrics.} Following the same practice as in \cite{fani2026dume}, we report the final results using normalized metrics relative to the performance each domain expert achieves in their respective domain (across all settings, the higher, the better). In particular, for both the causal language modeling settings, for each method, and for each domain score, we divide the perplexity score of the expert associated with that domain ($\Hat{p}_d$) by the perplexity score achieved by the method in that domain ($p_d$): $\Bar{p}_d = \frac{\Hat{p}_d}{p_d}$. For \reasoning, we adopt a similar approach, but in this case we invert the equation, given that in the \reasoning setting a higher (non-normalized) final score means a better performance: $\Bar{p}_d = \frac{p_d}{\Hat{p}_d}$. Finally, all our normalized scores are presented as percentages (so we multiply all the values by 100).
With these definitions, scores above 100 are possible because of randomness in the generation process, as detailed in the following section.

In the following section, we present the experimental results. 
In all experiments, the best and second best results (excluding the corresponding expert of each domain) are highlighted in \textbf{bold} and \underline{underlined}, respectively.

\subsection{Results}

\begin{table}[t]
    \centering
    \begin{minipage}{0.49\linewidth}
        \caption{Results in the \clm setting.}
        \label{tab:clm}
        \begin{adjustbox}{width=\linewidth}
            \begin{tabular}{c|ccccc|c}
                \toprule
                \textbf{Method} & \textbf{Coding} & \textbf{Math} & \textbf{Physics} & \textbf{History} & \textbf{Philosophy} & \textbf{Average} \\
                \midrule
                Coding expert     & 100.0& 76.1 & 84.9 & \underline{99.4} & 85.8 & 86.3 \\
                Math expert       & 93.5 & 100.0& 84.7 & 79.7 & 81.1 & 87.8 \\
                Physics expert    & 86.9 & 68.2 & 100.0& 75.1 & 86.0 & 85.3 \\
                History expert    & 74.2 & 49.3 & 74.6 & 100.0& 98.7 & 79.4 \\
                Philosophy expert & 75.6 & 51.4 & 75.5 & 98.1 & 100.0& 80.1 \\
                Experts average   & 90.9 & 72.2 & 88.4 & 93.5 & 94.4 & 87.9 \\
                \midrule
                Random router     & 84.9 & 67.0 & 84.1 & 89.1 & 90.3 & 83.1 \\
                MoDEM-small       & 97.0 & 95.9 & 94.2 & 98.5 & \textbf{102.0} & 97.6 \\
                MoDEM-large       & 97.4 & 95.5 & 97.3 & \underline{99.4} & \textbf{102.0} & 98.3 \\
                kNN router       & \textbf{99.9} & \underline{100.2} & 99.5 & \textbf{99.5} & \underline{100.7} & \textbf{100.0} \\
                \rowcolor{gray!15}
                \textbf{\method-avg (ours)}       & \textbf{99.9} & 94.4 & \underline{100.6} & 97.5 & 97.5 & 98.2 \\
                \rowcolor{gray!15}
                \textbf{\method-all (ours)}       & \underline{99.8} & \textbf{101.6} & \textbf{101.2} & 98.5 & 98.6 & \textbf{100.0} \\
                \rowcolor{gray!15}
                \textbf{\method (ours)}           & 99.1 & 94.6 & 99.5 & 98.1 & 98.6 & 98.2 \\
                \bottomrule
            \end{tabular}
        \end{adjustbox}
    \end{minipage}
    \hfill
    \begin{minipage}{0.44\linewidth}
        \caption{Results in the \clmII setting.}
        \label{tab:clm2}
        \centering
        \begin{adjustbox}{width=\linewidth}
            \begin{tabular}{c|cccc|c}
                \toprule
                \textbf{Method} & \textbf{Math} & \textbf{Biology} & \textbf{Legal} & \textbf{Dialogue} & \textbf{Average} \\
                \midrule
                Math expert       & 100.0 & 91.5 & 41.2 & 89.5 & 74.1 \\
                Biology expert    & 71.4 & 100.0 & 30.9 & 65.4 & 65.4 \\
                Legal expert      & 71.4 & 86.0 & 100.0 & 89.5 & 91.8 \\
                Dialogue expert   & 66.4 & 87.8 & 33.0 & 100.0 & 73.6 \\
                Experts average   & 79.8 & 93.5 & 49.7 & 89.5 & 77.5 \\
                \midrule
                Random router     & 75.7 & 88.3 & 45.1 & 87.2 & 73.5 \\
                MoDEM-small       & 93.5 & 89.0 & 86.2 & 84.2 & 86.5 \\
                MoDEM-large       & 97.5 & \underline{95.8} & 85.8 & 79.4 & 87.0 \\
                kNN router        & \underline{98.4} & \textbf{96.2} & \textbf{98.8} & \underline{98.8} & \textbf{97.9} \\
                \rowcolor{gray!15}
                \textbf{\method-avg (ours)}       & 88.1 & 90.3 & 82.7 & \textbf{99.4} & 90.8 \\
                \rowcolor{gray!15}
                \textbf{\method-all (ours)}       & 86.6 & 90.1 & 87.0 & \underline{98.8} & 92.0 \\
                \rowcolor{gray!15}
                \textbf{\method (ours)}           & \textbf{98.5} & \underline{95.8} & \underline{92.5} & 97.7 & \underline{95.3} \\
                \bottomrule
            \end{tabular}
        \end{adjustbox}
    \end{minipage}
\end{table}

\paragraph{Causal Language Modeling results.} \Cref{tab:clm,tab:clm2} present the results for the \clm and \clmII settings, respectively. For \textit{kNN router}, we report results for $\text{k}=10$ in the \clm setting, and $\text{k}=1$ in the \clmII setting; for \method, we report results with $k=100$ in the \clm setting, and $k=10$ for the \clmII setting. These values provided the best results for these methods.

As shown in the tables, our proposed methods are competitive with the baselines and, unlike \emph{kNN-router} and \emph{MoDEM}, do not require either an additional language model for embedding generation or the centralization of domain datasets. In particular, in \clm, they even surpass all of them in the Coding, Math, and Physics domains, and also in terms of average performance. It is worth noting that, in this setting, both \emph{kNN router} and our \method-all achieve an average performance of 100.0, meaning that, on average, they perform as well as each expert domain does in its own domain.

\begin{wraptable}{r}{0.5\textwidth}
    \caption{Results in the \reasoning setting.}
    \label{tab:reasoning}
    \centering
    \begin{adjustbox}{width=\linewidth}
        \begin{tabular}{c|cccc|c}
            \toprule
            \textbf{Method} & \textbf{Math} & \textbf{Multilingual} & \textbf{Coding} & \textbf{Instruction} & \textbf{Average} \\
            \midrule
            Math expert             & 100.0& 95.7 & 78.4 & 43.8 & 79.5 \\
            Multilingual expert     & 6.1  & 100.0& 73.1 & 21.2 & 50.1 \\
            Coding expert           & 11.7 & 99.1 & 100.0& 55.2 & 66.5 \\
            Instr. following expert & 28.9 & \underline{99.7} & 80.1 & 100.0& 77.2 \\
            Experts average         & 45.9 & 100.0 & 83.9 & 53.4 & 70.8 \\
            \midrule
            Random router           & 37.7 & 98.4 & 83.8 & 59.3 & 69.8 \\
            MoDEM-small             & 29.5 & 98.6 & 69.5 & 100.3 & 74.5 \\
            MoDEM-large             & 29.5 & 96.7 & 67.7 & 95.3 & 72.3 \\
            kNN router              & 95.8 & 99.5 & \textbf{96.3} & 98.7 & \underline{97.6} \\
            \rowcolor{gray!15}
            \textbf{\method-avg (ours)}   & \underline{97.3} & \textbf{99.9} & 85.6 & \textbf{101.2} & 96.0 \\
            \rowcolor{gray!15}
            \textbf{\method-all (ours)}   & 91.5 & 99.1 & 89.2 & 100.1 & 95.0 \\
            \rowcolor{gray!15}
            \textbf{\method (ours)}       & \textbf{98.4} & \textbf{99.9} & \underline{94.5} & \underline{100.6} & \textbf{98.4} \\
            \bottomrule
        \end{tabular}
    \end{adjustbox}
\end{wraptable}

\paragraph{\reasoning.} \Cref{tab:reasoning} shows the results in the \reasoning setting. For \textit{kNN router}, we report results for $\text{k}=1$; for \method, we report results with $k=10$, as these values provided the best results for these methods.

\begin{figure*}[t]
    \centering
    \begin{minipage}{0.32\linewidth}
        \centering
        \includegraphics[width=\linewidth]{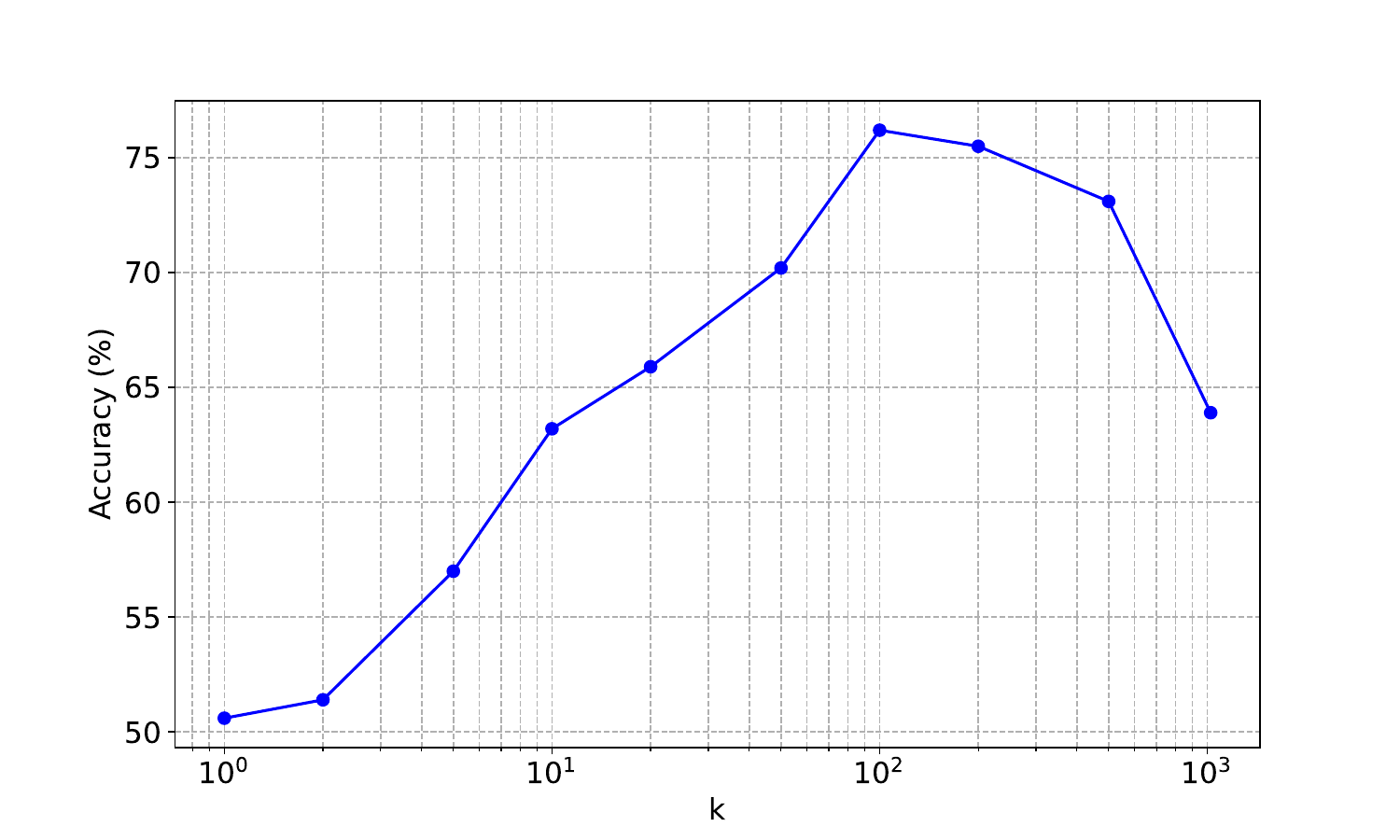}
    \end{minipage}
    \hfill
    \begin{minipage}{0.32\linewidth}
        \centering
        \includegraphics[width=\linewidth]{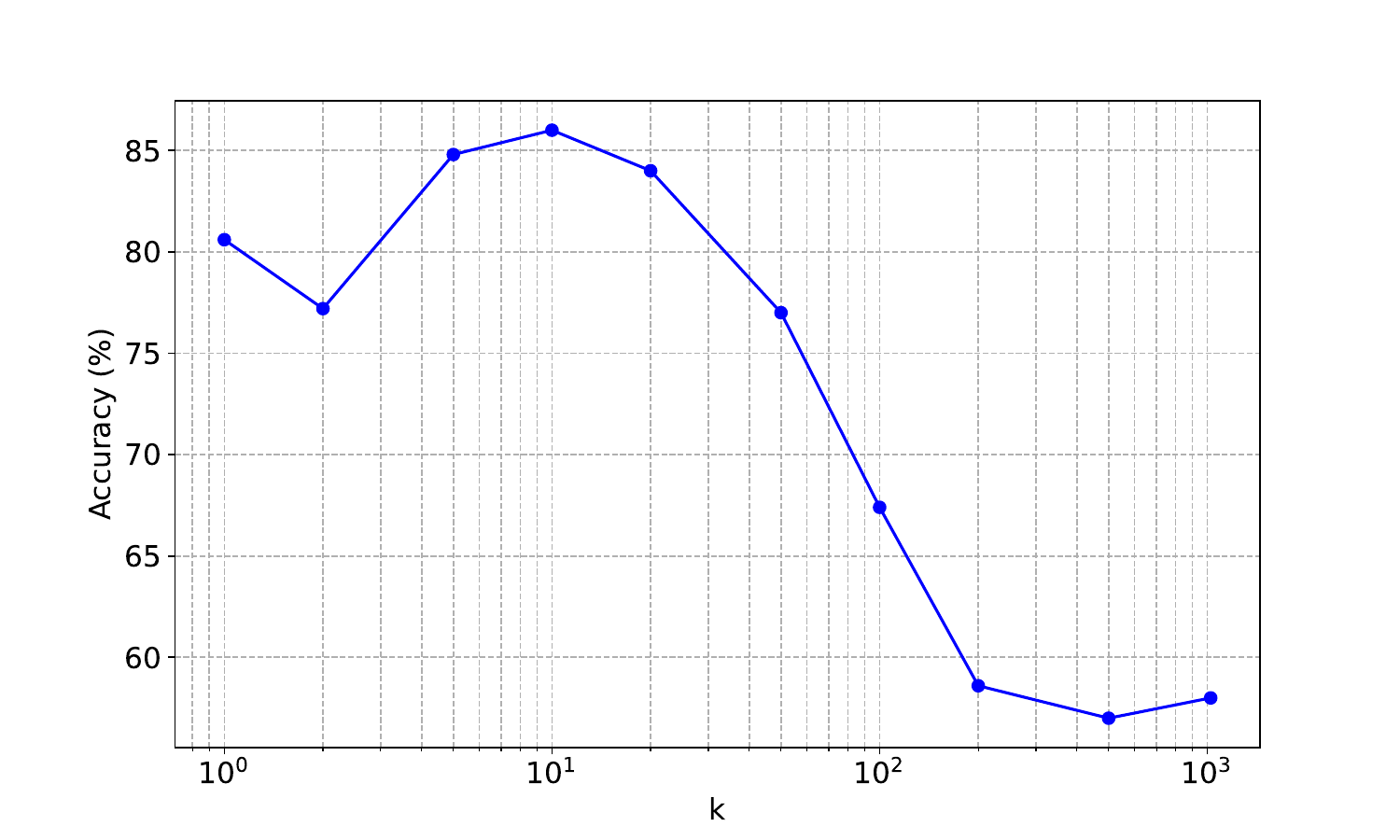}
    \end{minipage}
    \hfill
    \begin{minipage}{0.32\linewidth}
        \centering
        \includegraphics[width=\linewidth]{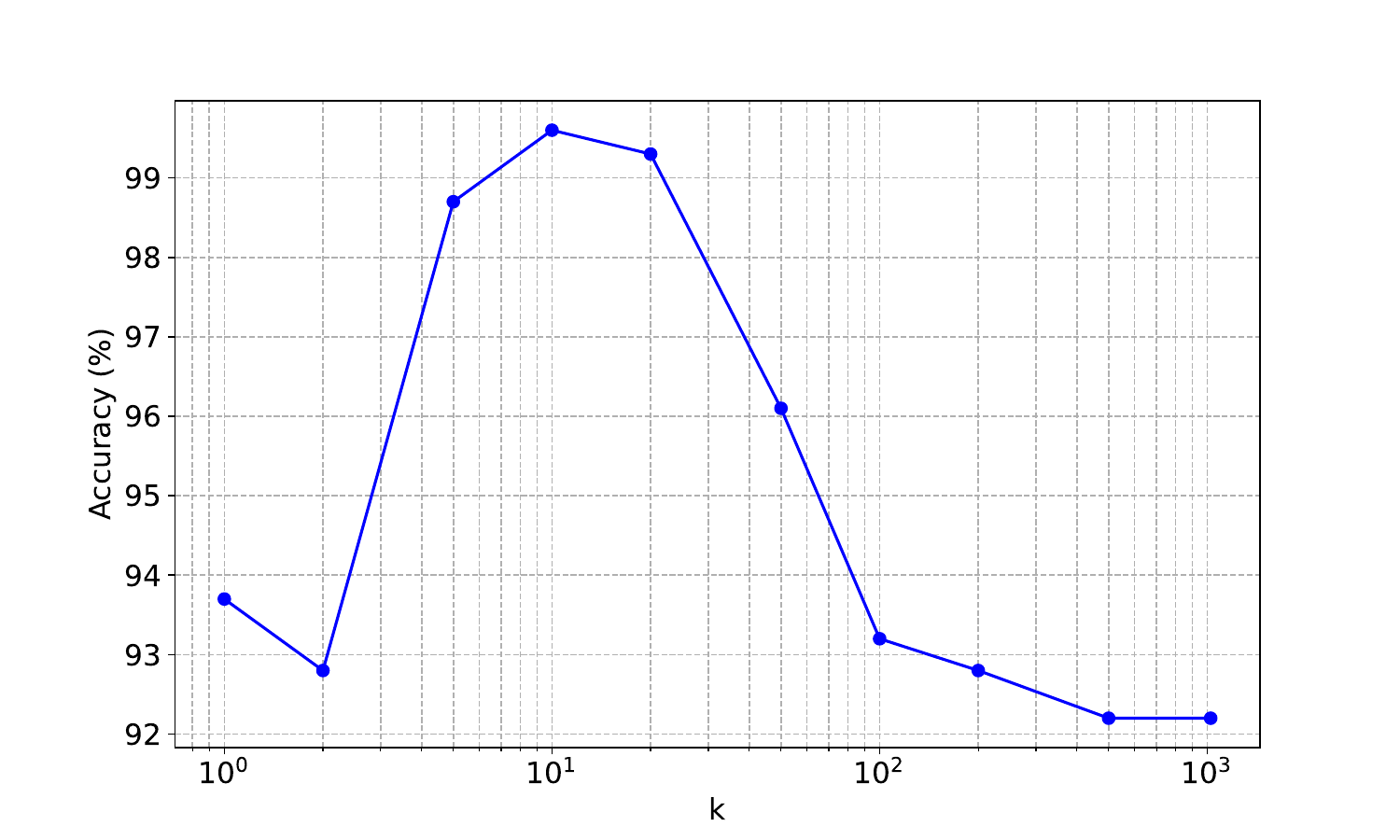}
    \end{minipage}
    \caption{\method routing accuracy (\%) with various values of $k$ for the top-k smallest entropy token filtering, on the \clm (left), \clmII (center) and \reasoning (right) settings.}
    \label{fig:entropyk}
\end{figure*}

In this setting, \method achieves the best average performance and the best or second-best performance across all single domains. In particular, it achieves a normalized average performance of 98.4. Comparably, \emph{kNN router}, the second-best performing method in terms of average performance, achieves 97.6. Notably, some of the scores exceed 100, indicating performance better than that of the domain expert in their respective domains. In principle, this should not be possible, since regardless of the router's decisions, only one of the original domain experts handles each sample. However, given that the text has a random component (we use a temperature of 0.7 for all text generation), this is actually possible by chance when the routing accuracy, \ie, the percentage of samples correctly routed to their corresponding expert, approaches 100.

In the next section, we analyze the routing accuracy of \method for various values of $k$.

\paragraph{On the Sample Route Selector routing mechanism of \method.}
\label{sec:kentropy}

The Sample Route Selector in \method filters tokens based on entropy. In particular, only the $k$ tokens, whose softmax predictions are among the $k$ smallest, are used to cast a final vote to determine the routing expert. This section answers the following question: \emph{Why not allow all the tokens to participate in majority voting?} In the previous sections, we have answered this question empirically, by showing that \method-all performs worse than \method in all three settings in terms of the average score. The motivation is that, when constructing the Token Router with ridge regression, the same (common) tokens can have different domain labels, as they likely appear in different domains. For instance, it is very likely that the token ``the'' will appear in all domains. However, when solving the RLS problem, having the same input vector repeated with different labels will yield uncertain predictions. For example, if there are only two domains, and the token ``the'' appears the same number of times in each domain, the probability score of the Token Router for ``the'' would be the vector $(0.5, 0.5)$. Therefore, the motivation for filtering based on entropy is that the tokens for which the Token Router is more uncertain are likely the ones that appear with equivalent frequency across all domains. Such tokens are therefore not discriminative and should be excluded from the decision-making process, as they will have a higher entropy score.

\Cref{fig:entropyk} confirms our intuition. In this figure, we present the routing accuracy (\ i.e., the percentage of prompts correctly routed to their corresponding domains) in all three settings with various values of $k$. As can be observed, the three settings follow the same trend. When only a bunch of tokens participate in the majority voting, the signal from each prompt is too shallow and \method provides a smaller routing accuracy. Similarly, when too many tokens participate, they introduce noise into the selection of the routing domain, since tokens with uncertain predictions are allowed to vote alongside those with more confidence. The best solution is always to allow a sufficiently large pool of confident tokens to participate, while excluding the uncertain ones.

\section{Conclusion}
\label{sec:conclusion}

As the number and diversity of available LLMs continue to grow, effective routing becomes increasingly important for selecting the most suitable model for each query. Existing routing methods have largely focused on cost–performance trade-offs among generalist models. Recently proposed routers also consider domain accuracy, assigning queries to specialist models based on their domain relevance. However, existing work often relies on language-model-based classifiers or embedding models, which increase routing cost and typically require access to domain datasets for training, raising privacy concerns. These limitations underscore the need for lightweight, adaptable routing methods for expert LLM selection. Despite being a linear router, \method achieves performance comparable to other baselines across all settings and even surpasses the second-best baseline, \emph{k-NN router}, in the \reasoning setting, achieving a normalized performance of 98.4\%. Moreover, in \clm, \method-all performs, on average, on par with expert domain models evaluated on their own domains. Together, these results highlight the strengths of \method and its variants as a cheap and fast alternative for inference routing.

While \method is lightweight and efficient thanks to its linear construction, it is also less expressive than stronger LLM-based routers; as a result, it may be less effective on queries that require richer semantic understanding or complex decision boundaries. This trade-off motivates several directions for future work. First, the current ridge-regression formulation could be extended to kernel ridge regression to capture non-linear structure while retaining much of the method's analytical simplicity. Second, it would be valuable to evaluate and adapt the router for more complex reasoning tasks, where domain relevance alone may be insufficient, and routing may need to account for multi-step reasoning requirements. Third, future versions of the router could explicitly incorporate system-level costs into the routing objective, including not only predictive performance but also computation, latency, and memory usage, enabling more practical deployment in resource-constrained settings.

\FloatBarrier

\bibliography{references}
\bibliographystyle{plainnat}

\FloatBarrier

\appendix
\section*{\LARGE Appendix}

\begin{figure}[ht!]
    \centering
    \begin{minipage}[t]{0.38\linewidth}
        \centering
        \includegraphics[width=\linewidth]{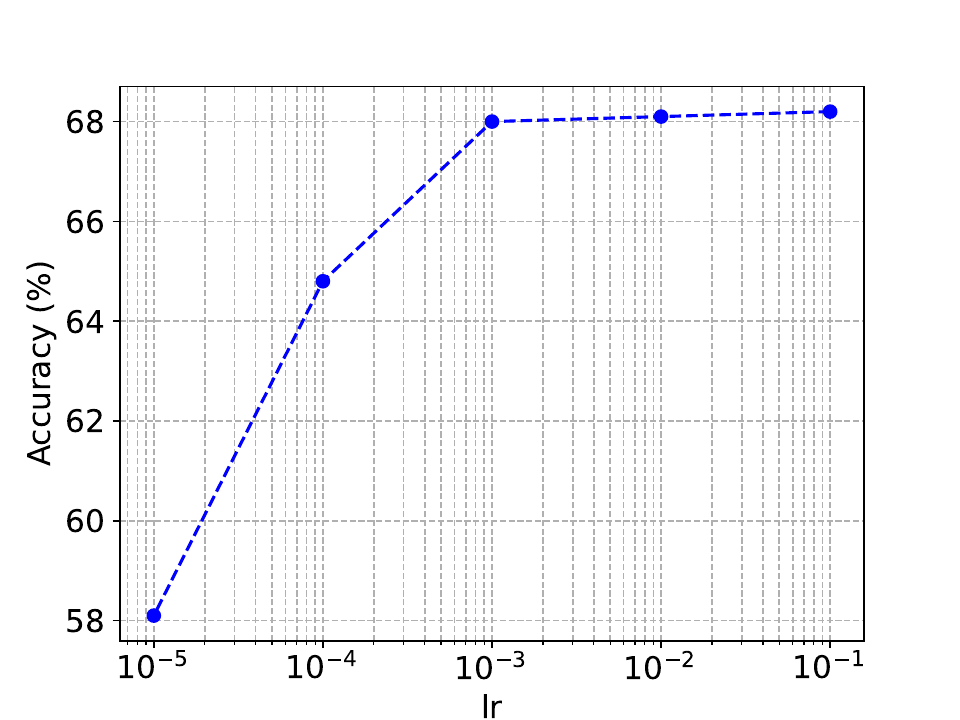}
    \end{minipage}
    \begin{minipage}[t]{0.38\linewidth}
        \centering
        \includegraphics[width=\linewidth]{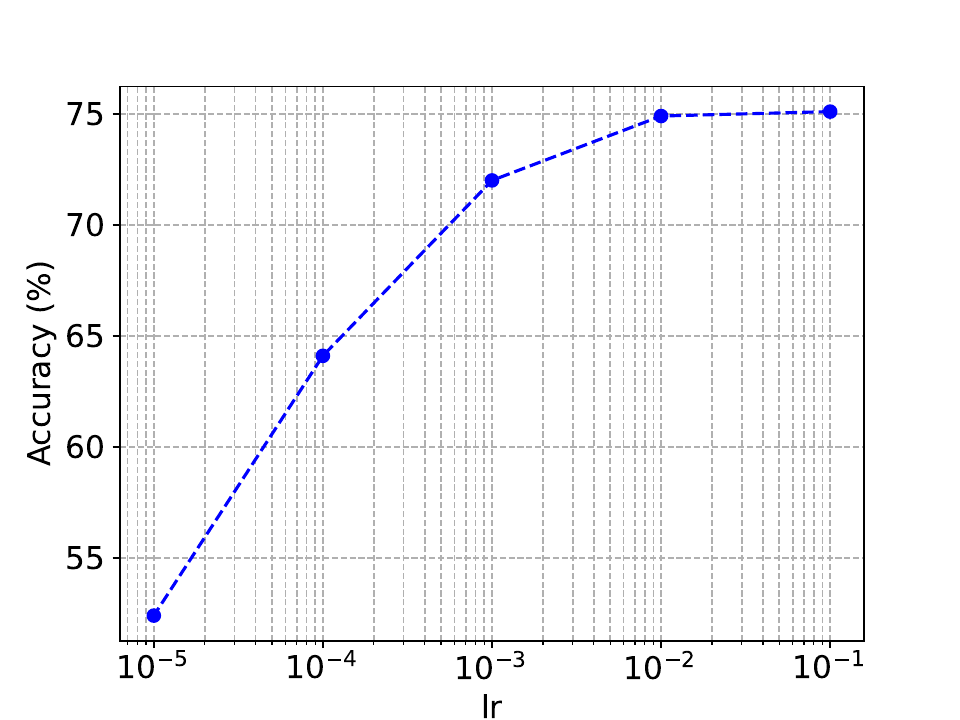}
    \end{minipage}
    \caption{Final accuracy (\%) of the MoDEM router with various values for the learning rate (\textit{lr}) in the \clm setting: small router (left) and large router (right).}
    \label{fig:modemlrclm}
\end{figure}

\begin{figure}[ht!]
    \centering
    \begin{minipage}[t]{0.38\linewidth}
        \centering
        \includegraphics[width=\linewidth]{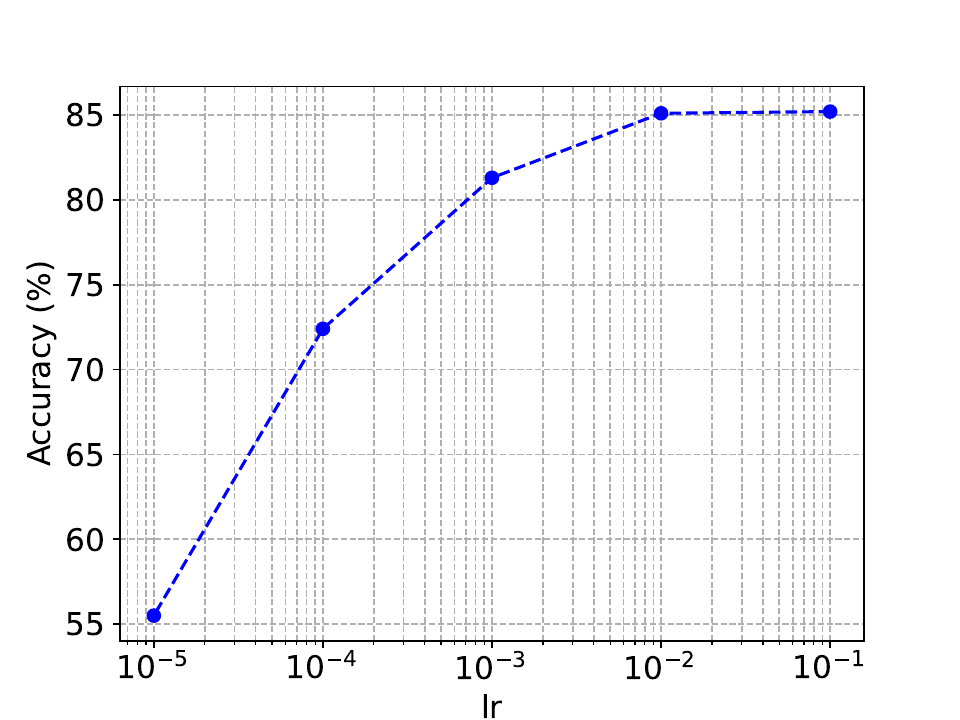}
    \end{minipage}
    \begin{minipage}[t]{0.38\linewidth}
        \centering
        \includegraphics[width=\linewidth]{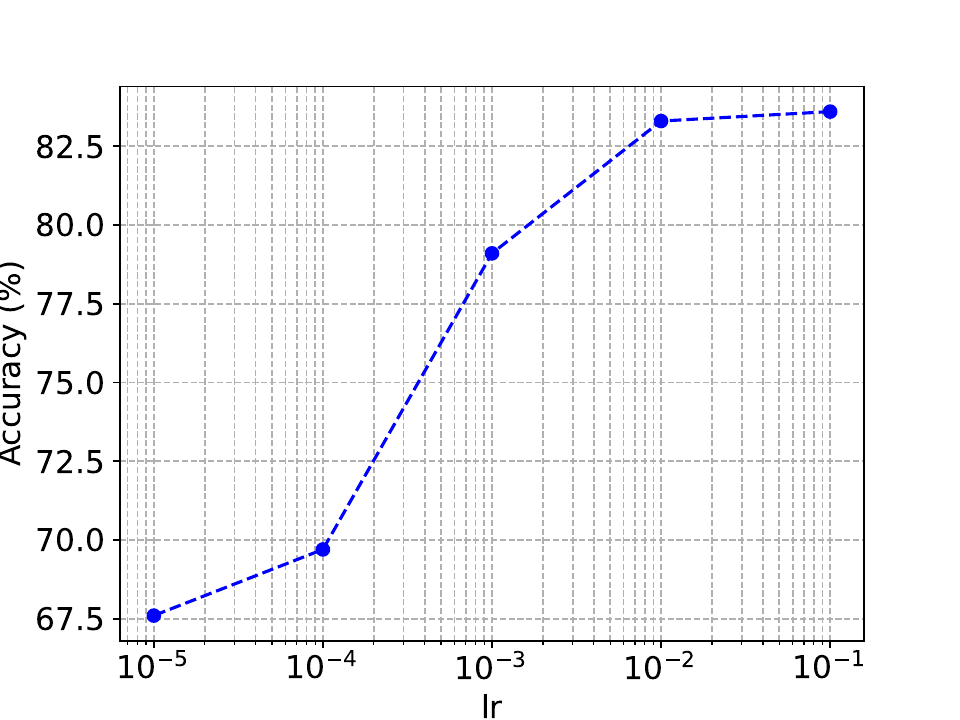}
    \end{minipage}
    \caption{Final accuracy (\%) of the MoDEM router with various values for the learning rate (\textit{lr}) in the \reasoning setting: small router (left) and large router (right).}
    \label{fig:modemlrreasoning}
\end{figure}

\section{Details on MoDEM router training}

We have trained the MoDEM routers for both \clm and \reasoning using a merged dataset across all domains, in the corresponding settings. For each domain, we fixed the number of samples at 1750 to avoid class imbalance. In the \reasoning setting, we trained both the small and large MoDEM routers for 100 epochs. Similarly, we trained the small MoDEM in the \clm setting for 100 epochs, whereas we found it sufficient to train the large MoDEM for only 10 epochs, as the best validation routing accuracy plateaued after a few epochs. For both the large and small MoDEM, in both settings, we trained MoDEM with a batch size of 16, clipped the gradient norms to 1, and a cosine annealing scheduler with a warmup.

\Cref{fig:modemlrclm,fig:modemlrreasoning} show the final test routing accuracies in both the \clm and \reasoning settings, respectively, with $lr \in {10^{-5}, 10^{-4}, 10^{-3}, 10^{-2}, 10^{-1}}$. In all cases, we eventually selected the MoDEM router trained with the largest learning rate, \ie $lr = 0.1$, as it provided the best test routing accuracy.

\end{document}